%% file: main.tex
\documentclass[runningheads]{llncs}

\usepackage{eccv}

\usepackage{eccvabbrv}

\usepackage{graphicx}
\usepackage{booktabs}

\usepackage[accsupp]{axessibility}  

\usepackage{verbatim}
\usepackage{xcolor,colortbl}
\usepackage{soul}
\usepackage{multirow}

\definecolor{LightGray}{gray}{0.9}
\definecolor{Gray}{gray}{0.8}
\definecolor{LightCyan}{rgb}{0.88,1,1}

\makeatletter
\def\@fnsymbol#1{\ensuremath{\ifcase#1\or *\or \dagger\or \ddagger\or
   \mathsection\or \mathparagraph\or \|\or **\or \dagger\dagger
   \or \ddagger\ddagger \else\@ctrerr\fi}}
\makeatother

\def\ourmethod{DPaRL}

\usepackage{hyperref}

\usepackage{orcidlink}

\begin{document}

\title{Open-World Dynamic Prompt and \\Continual Visual Representation Learning}

\titlerunning{DPaRL for Open-World Continual Learning}


\author{Youngeun Kim\inst{1,}\thanks{Equal contribution.}\thanks{Work conducted during an internship at Amazon.}, \hspace{0.1cm} Jun Fang\inst{2,3, *}\thanks{Corresponding author.}, \hspace{0.1cm} Qin Zhang$^2$, \hspace{0.1cm} Zhaowei Cai$^3$, \\ Yantao Shen$^2$, \hspace{0.1cm} 
Rahul Duggal$^2$, \hspace{0.1cm} Dripta S. Raychaudhuri$^2$, \\ Zhuowen Tu$^2$, \hspace{0.1cm} Yifan Xing$^2$, \hspace{0.1cm} Onkar Dabeer$^2$ 
}

\authorrunning{Y.~Kim et al.}


\institute{Yale University \and AWS AI Labs \and Amazon AGI
\\ \email{youngeun.kim@yale.edu  }  \email{\{junfa, qzaamz, zhaoweic, yantaos, dugrahul, driptarc, ztu, yifax, onkardab\}@amazon.com}
}

\maketitle

\begin{abstract}
  The open world is inherently dynamic, characterized by ever-evolving concepts and distributions. Continual learning (CL) in this dynamic open-world environment presents a significant challenge in effectively generalizing to unseen test-time classes. To address this challenge, we introduce a new practical CL setting tailored for open-world visual representation learning. In this setting, subsequent data streams systematically introduce novel classes that are disjoint from those seen in previous training phases, while also remaining distinct from the unseen test classes. In response, we present \textbf{D}ynamic \textbf{P}rompt \textbf{a}nd \textbf{R}epresentation \textbf{L}earner (\textbf{DPaRL}), a simple yet effective Prompt-based CL (PCL) method. Our DPaRL learns to generate \textit{dynamic} prompts for inference, as opposed to relying on a \textit{static} prompt pool in previous PCL methods. In addition, DPaRL jointly learns dynamic prompt generation and discriminative representation at each training stage whereas prior PCL methods only refine the prompt learning throughout the process. Our experimental results demonstrate the superiority of our approach, surpassing state-of-the-art methods on well-established open-world image retrieval benchmarks by an average of 4.7\% improvement in Recall@1 performance.  
  \keywords{Dynamic Prompt Generation \and Continual Learning \and Open-World Visual Representation Learning}
\end{abstract}

\input{sections/1_introduction}
\input{sections/2_related_work}
\input{sections/3_method}
\input{sections/4_experiments}

\input{sections/5_conclusion}


%
%
\bibliographystyle{splncs04}
\bibliography{main}

\clearpage

\input{supp_material}

\end{document}

%% file: sections/1_introduction.tex
\section{Introduction} \label{sec:intro}

\begin{figure}[tb]
\begin{center}
\centering
\def\arraystretch{0.95}
\begin{tabular}{@{\hskip 0.0\linewidth}c@{\hskip 0.0\linewidth}c@{\hskip 0.0\linewidth}c@{\hskip 0.0\linewidth}c}
  \includegraphics[width=0.22\linewidth]{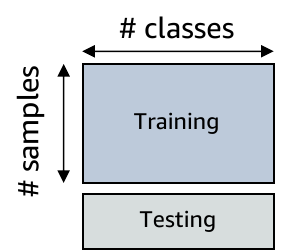} 
& \includegraphics[width=0.22\linewidth]{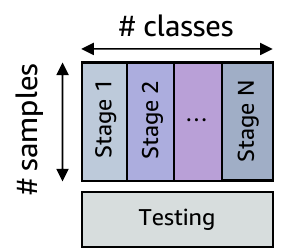}
& \includegraphics[width=0.26\linewidth]{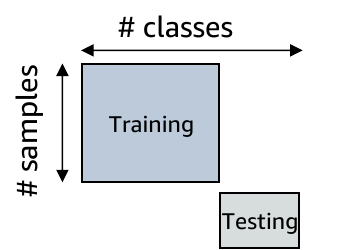} 
& \includegraphics[width=0.26\linewidth]{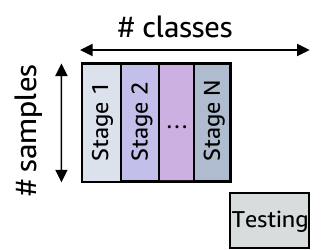}
\\
{\hspace{5mm}(a)}  & {\hspace{4mm}(b)} & {(c)}  & {(d)}
\end{tabular}
\end{center}
\caption{ 
Illustration of the distinctions between our problem setting and traditional settings. In our work, we aim to address the problem where training splits have no class overlaps.
(a) Closed-world setting: both training and testing classes are identical.
(b) Continual learning (CL) in closed-world setting: training classes are split into multiple divisions and are introduced through various CL stages.
(c) Open-world setting: training and testing classes remain separate. The aim is to learn a robust representation from training classes that generalize to unseen classes.
(d) \textbf{Open-World Continual Representation Learning (Our Problem Setting)}: 
continual learning tailored for an open-world scenario. We sequentially introduce training classes over multiple CL stages, ensuring they remain distinct from test-time unseen classes.
}
\label{fig:problem_setting}
\end{figure}

Continual learning (CL) without catastrophic forgetting is a challenging task in practice as retraining with old data is prohibitively expensive when data volume grows \cite{de2021continual,mai2022online,hadsell2020embracing}. It is particularly challenging for real-world applications in the \textit{open-world}~\cite{an2023unicom,cao2022pss}, where new concepts keep emerging and distribution shifts are inevitable. Recent studies on Prompt-based Continual Learning (PCL) ~\cite{wang2022learning, wang2022dualprompt, smith2023coda} demonstrated success in mitigating catastrophic forgetting without access to any past data. However, they heavily focus on the \textit{closed-world} setting where test-time categories come from those encountered in various training stages. 

Whereas in practice, the world is dynamic and concepts evolve over time and new categories are often introduced after a model is trained. Consider, for instance, an image retrieval system for clothing brands in online e-commerce. During training, one would gather images of all existing clothing brands available to train a model. However, after deployment, the system encounters new brands. The new data collected for system upgrade, constituting a new training stage, will likely include brands that are disjoint to those in older stages. Furthermore, the system will constantly face new brands at test-time that are disjoint to any of those encountered in the previous training stages and is expected to generalize well to those new categories. 

Thus, to comprehensively evaluate the ability of a model in retaining semantic information from continuous streams of data for generalization to new concepts at test-time for practical applications, we propose a new continual \textit{open-world} representation learning setting, as illustrated in Figure~\ref{fig:problem_setting}(d). In this setting, different data streams comprise distinct classes, and test-time classes are disjoint to those encountered in all training phases.
Our setting is notably more challenging than the previous closed-world CL scenario. It closely mirrors real-world demands on recognition systems, where, during any evaluation stage, the test set consists of unseen entities to the training process up to that point. This setup reflects the need for practical recognition systems to encode general semantic information from training-time classes and apply this knowledge to new, unseen concepts without access to older training data during system updates.

We start with evaluating state-of-the-art PCL methods~\cite{wang2022learning, wang2022dualprompt, smith2023coda} in our proposed setting and find that they fail to generalize with the absence of past data. For example, in our experiments, the gap between the paragon of standard training with all data in one phase and the best PCL baseline (CodaPrompt~\cite{smith2023coda}) shows up to $20\%$ gap in Recall@1 (Table~\ref{table:exp:comparison}), significantly larger than the $5\%$ gap in the closed-world setting~\cite{smith2023coda}, which suggests that prior PCL methods encounter difficulties and are less effective in the open-world setting. 

\begin{figure*}[t]
\begin{center}
\centering
\def\arraystretch{0.95}
\begin{tabular}{@{\hskip 0.01\linewidth}c@{\hskip 0.03\linewidth}c@{}c}
\includegraphics[width=0.46\linewidth]{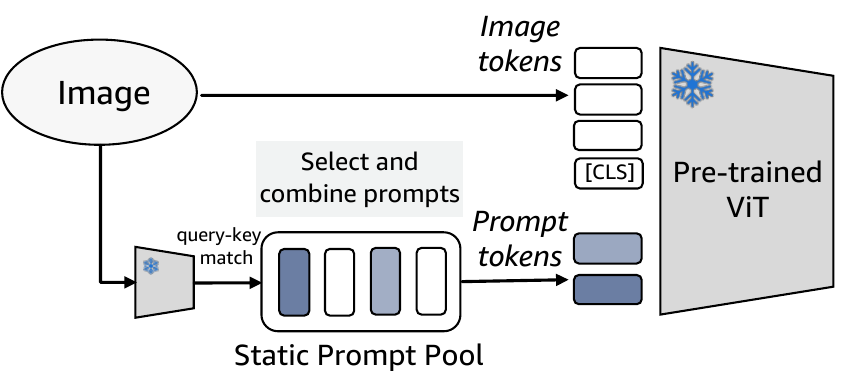} 
&
\includegraphics[width=0.46\linewidth]{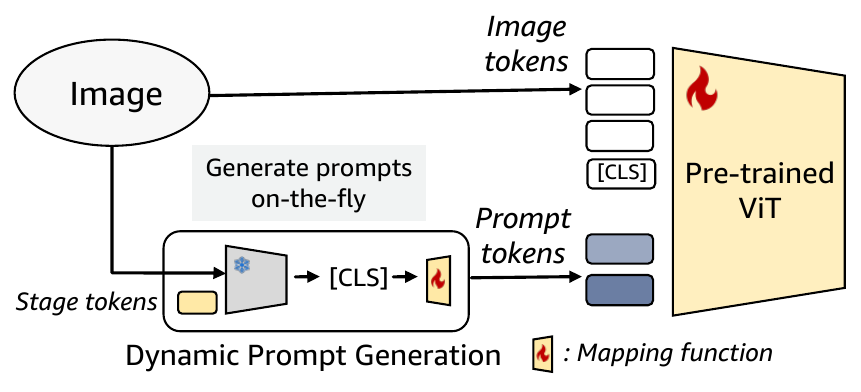}
\\
{(a) Previous method}& {(b) Our work}
\end{tabular}
\end{center}
\caption{ 
Prior PCL methods combine prompts from a \textit{static} prompt pool trained on the training class distribution, leading to a loss of generalization capability when facing unseen classes during test time. 
Our work introduces a Dynamic Prompt Generation network that generates \textit{dynamic} prompts on the fly by integrating a given image with stage tokens, followed by a specialized mapping function and adjustable discriminative representation backbone weights, providing generalizable prompts for unseen testing classes, distinguishing it from prior PCL methods. 
}
\label{fig:ours_vs_previous}
\end{figure*}

To tackle this challenging yet practical open-world CL setting, we introduce a simple yet effective method called Dynamic Prompt and Representation Learner (DPaRL). Unlike previous prompt pool-based PCL methods~\cite{wang2022learning, wang2022dualprompt, smith2023coda} that depend on a static prompt pool (Figure~\ref{fig:ours_vs_previous}(a)) for test-time predictions. Notably, DPaRL trains an innovative dynamic prompt generation network jointly with the discriminative representation backbone model (Figure~\ref{fig:ours_vs_previous}(b)), departing from the prompt-only learning paradigm in prior PCL methods.

Our key insight lies in that simply combining the fixed prompts learnt over training classes does not generalize well to unseen test-time classes in the practical open-world setting. On the other hand, the on-the-fly dynamically generated prompts in our method exhibit a stronger capability of capturing the diverse semantics spread across the numerous divisions of classes observed during the different training stages of continual learning, improving generalization to test-time unseen classes. This is achieved through the proposed \textit{dynamic} prompt generation network, which possesses stronger representation power with a learnable mapping function compared to the traditional approach of combining multiple prompts in a \textit{static} prompt pool. Additionally, unlike prior PCL methods, DPaRL jointly updates the dynamic prompt generation process and the discriminative representation backbone weights, facilitating maximized interaction and comparison between old concepts and newer ones to encapsulate diverse semantics that aid generalization to unseen test-time classes.

Consequently, over the four open-world image retrieval benchmarks on cars, online catalogs, products, and species images where the task is to search for the most similar images belonging to the same class as the query, via a distance measurement in an embedding space, DPaRL achieves significant accuracy boosts. It improves the average Recall@1 performance by absolute 4.7\% and 9.1\% compared to state-of-the-art continual learning methods, from 76.1\% to 80.8\% over 10 CL stages and from 68.0\% to 77.1\% over 100 CL stages, respectively.

In summary, our contributions are three folds:
\begin{enumerate}
    \item We establish a new practical setting on continual visual representation learning in the open-world.
    \item We introduce a simple yet powerful method, Dynamic Prompt and Representation Learner (DPaRL), which dynamically generates prompts while effectively updating the discriminative representation backbone. This enhancement improves generalization for unseen open-world classes at test time.
    \item We outperform state-of-the-art continual learning methods in the proposed practical setting, both rehearsal-free and rehearsal-based.
\end{enumerate}

%% file: sections/2_related_work.tex
\section{Related Work}

\noindent\textbf{Continual learning.} 
Continual learning (CL) has been recognized for decades as a crucial research domain, focusing on the ongoing adaptation of models over time. Early works primarily centered around \textit{regularization-based} methods in an effort to address catastrophic forgetting \cite{aljundi2018memory,li2017learning,zenke2017continual} by introducing regularization between its current and previously learned parameters. However, they often underperform when faced with complex datasets.

A separate line of work, \textit{architecture-based} approach, involves expanding the network architecture in response to new learning phases \cite{li2019learn,rusu2016progressive,yoon2017lifelong,serra2018overcoming,rao2019continual}. While these methods generally outperform regularization-based counterparts, they do so at the cost of introducing additional parameters.
Recent advancements include \textit{rehearsal-based} methods that incorporate a memory component for storing past data \cite{chaudhry2018efficient,chaudhry2019tiny,hayes2019memory,buzzega2020dark,rebuffi2017icarl}. This archived data aids in training newer stages, typically yielding superior performance compared to other CL techniques. Yet, this data retention can introduce privacy concerns and significant memory overhead.

Given these constraints, \textit{rehearsal-free} strategies are gaining attention in the community. These aim to mitigate catastrophic forgetting without relying on historical data. A prominent method within this category uses model inversion to produce rehearsal images \cite{choi2021dual,gao2022r,yin2020dreaming,smith2021always}. 
However, model inversion is computationally intensive and time-consuming. Additionally, these methods lag significantly in performance when compared to their rehearsal-based counterparts.

Recent developments have witnessed the rise of \textit{prompt-based} approaches~\cite{wang2022learning, wang2022dualprompt, smith2023coda, zhou2023revisiting, tang2023prompt, wang2023hierarchical, razdaibiedina2023progressive}. These strategies offer robust protection against catastrophic forgetting by learning a small set of parameters, or prompts, instead of directly training all parameters. For instance, L2P \cite{wang2022learning} employs a pool of prompts, selecting suitable ones for insertion based on input data clustering. Building 
on this concept, DualPrompt \cite{wang2022dualprompt} presents general prompts designed to encapsulate shared knowledge across various learning stages. Advancing this further, the state-of-the-art method CodaPrompt \cite{smith2023coda} facilitates end-to-end training by incorporating an attention mechanism for prompt combination within a pool. However, during inference, the above works rely on the combination of prompts from a static pool learned from training to feed into a frozen backbone model, which has limitations in cases where testing data has distribution shifts or disjoint classes from the training data. This motivates us to design a Dynamic Prompt and Representation Learner (DPaRL) to address these drawbacks.

\subsubsection{Open-World Visual Representation Learning.}
In open-world visual representation learning, the model learns  discriminative representations that align distances between representations with their semantic similarities, and uses these knowledge for testing on unseen classes. Previous works in this domain have primarily focused on designing loss functions \cite{movshovitz2017no,deng2019arcface,kim2020proxy,zhang2024threshold} for enhanced accuracy. Recently, several studies~\cite{dosovitskiy2020image,an2023unicom} have demonstrated the effectiveness of utilizing pre-trained visual foundation models and fine-tuning them to achieve significant performance improvements.
While open-world visual representation learning has been under investigation for an extended period, there has been limited exploration within the continual learning (CL) framework for this task. \cite{zhao2021continual} investigated biometric datasets using a regularization-based CL approach. However, the investigation into state-of-the-art prompt-based CL methods remains missing. Furthermore, as their studies primarily centered on biometric data, there is a pressing need to explore established continual learning benchmarks specific to open-world visual representation learning in more natural settings.

%% file: sections/3_method.tex
\section{Methodology}

In this section, we introduce our open-world Continual Learning (CL) setting and motivate our Dynamic Prompt and Representation Learner (DPaRL).

\subsection{Problem Setting} 

Two fundamental visual representation learning settings are the \textit{closed-world} and \textit{open-world} paradigms. In a \textit{closed-world} setting (as depicted in Figure \ref{fig:problem_setting}(a)), the training and testing data classes are entirely identical. Conversely, the \textit{open-world} setting (illustrated in Figure \ref{fig:problem_setting}(c)) presents a more practical scenario. Here, training and testing classes are entirely distinct, thus requiring the model to learn representations that generalize to unseen concepts.

When faced with continuous streams of data for visual recognition system updates, built upon the \textit{closed-world} paradigm, Figure \ref{fig:problem_setting}(b) showcases the well-known continual learning setting for image recognition.
In this work, we introduce a new practical setting: \textit{Open-World Continual Representation Learning} as shown in Figure~\ref{fig:problem_setting}(d). Instead of testing on seen classes of the various training stages thus far, the testing classes are entirely disjoint to those faced in different training stages, necessitating the model to not only learn continually but also generalize to new concepts, making it a more dynamic and challenging setting.

\subsection{Prompt-based Continual Learning Paradigm} 

Prompt-based Continual Learning (PCL) methods~\cite{wang2022learning, wang2022dualprompt, smith2023coda} use a pre-trained Vision Transformer (ViT) as a discriminative backbone for \textit{closed-world} image classification, as shown in the Figure~\ref{fig:ours_vs_previous}(a). These methods create a prompt pool with multiple prompt tokens, updating only the learnable parameters in this pool during training. In inference, the learned prompt pool is \textit{static}, and PCL methods select tokens from this pool to feed into multiple ViT backbone layers for predictions. Our method adopts this PCL paradigm but adapts to the \textit{open-world} representation learning setting with a prototypical metric learning loss~\cite{deng2019arcface}. Additionally, we introduce a \textit{dynamic} prompt generation (DPG) network to replace the static prompt pool for effective joint learning with the discriminative representation backbone model, as detailed below.

\begin{figure*}[t]
\begin{center}
\centering
\def\arraystretch{0.5}
\begin{tabular}
{@{\hskip -0.01\linewidth}c@{\hskip 0.03\linewidth}c@{}c}
\includegraphics[width=0.95\linewidth] 
{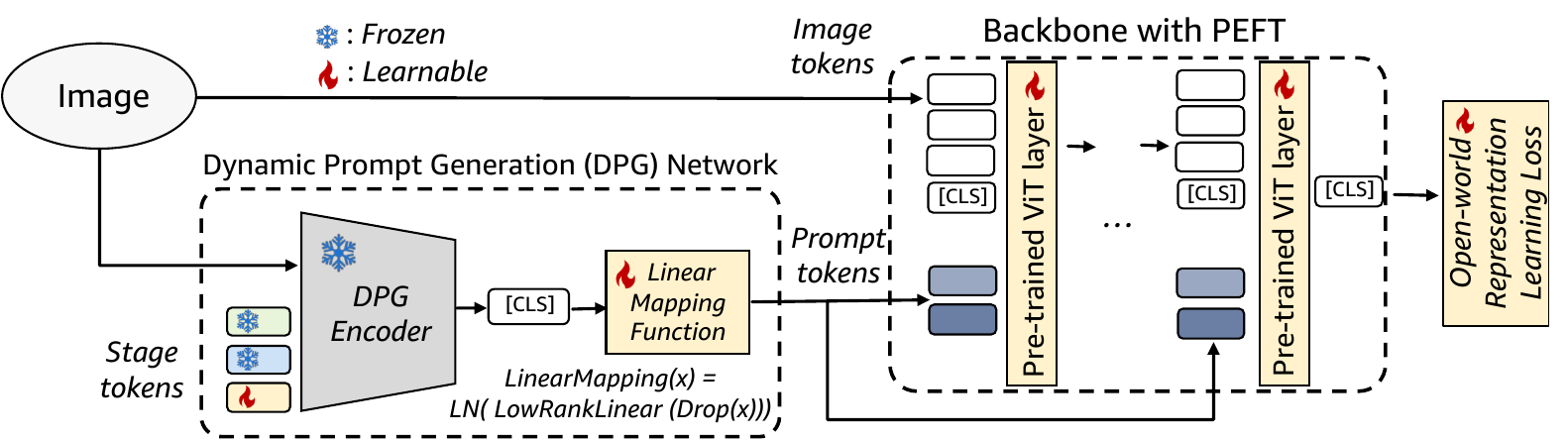}
\end{tabular}
\end{center}
\caption{ 
Illustration of our proposed method: Dynamic Prompt and Representation Leaner (\ourmethod). We dynamically generate prompt tokens on the fly from a Dynamic Prompt Generation (DPG) network by integrating information from stage tokens and image tokens. The [CLS] token from DPG is converted to prompt tokens via a low-rank linear mapping function. The generated prompt tokens are added to the backbone ViT and trained with a loss function. The learnable parameters include the current stage token, weights in mapping function, weights in backbone, and weights in loss function. 
}
\label{fig:method}
\end{figure*}

\subsection{Dynamic Prompt and Representation Learner (\ourmethod)}\label{sec:DPG}

\subsubsection{Overall Pipeline.} 
In the open-world setting where training and testing classes are disjoint, existing prompt pool designs in PCL \cite{wang2022learning, wang2022dualprompt, smith2023coda} exhibit limitations, i.e., limited separation between intra- and inter- testing class distance distributions as shown in Figure~\ref{fig:hist_l2_dist} and large performance gap from paragon in Table~\ref{table:exp:comparison}. To mitigate, we first propose dynamically generating prompts by blending the given image information and continual learning stage's information through a dynamic prompt generation (DPG) network. It aims to provide better representation power compared to the prompt pool design, which combines a fixed prompt pool trained from the training class distribution. Moreover, we enhance the capability of the discriminative representation backbone model to more effectively generalize to the open-world concepts through a joint dynamic prompt and representation learning paradigm. We refer to Figure~\ref{fig:ours_vs_previous}, \ref{fig:method} for detailed design differences between our approach and prior prompt pool-based methods.

\subsubsection{Dynamic Prompt Generation Network.} 
In order to retain information from each CL stage of our Dynamic Prompt Generation (DPG) process, we introduce stage tokens, denoted as $S$. During training stage $t$, we train a stage token $S_{t}$ while freezing the previous stage tokens $S_{t - (q - 1)} \sim S_{t-1}$ in a first-in-first-out (FIFO) queue manner, where $q$ is the max queue size. Guided by the ablation study detailed in Section~\ref{sec:exp:stage_token}, we typically set $q = 5$ using a FIFO order to aim for both high accuracy and scalability.
This ensures that the knowledge from the previous stages remains unchanged, and the total number of stage tokens limits by the queue size without linearly scaling with stage numbers.
Within the deep neural network such as ViT, the self-attention block facilitates the integration of information between the stage-wise tokens and the instance-wise image tokens.

As illustrated in Figure~\ref{fig:method}, the DPG produces a [CLS] token to dynamically acquire task-specific high-level information for prior knowledge of class-level features. However, it differs in size from the prompt tokens required by the discriminative backbone model (right side of Figure~\ref{fig:method}). To address this, we introduce a mapping function between the [CLS] token and the prompt tokens. 

A straightforward approach is to employ a single linear layer for dimensionality transformation. However, such a method introduces an excessive number of additional parameters, and the resulting over-parameterization can lead to overfitting of the highly compressed [CLS] token information.   
To tackle this issue, we impose a constraint on the weight parameter $W=AB^T \in \mathbb{R}^{C_{in} \times C_{out}}$, where $A \in \mathbb{R}^{C_{in} \times R}$, $B \in \mathbb{R}^{C_{out} \times R}$, ensuring a maximum rank of $R< min(C_{in}, C_{out})$. This constraint is inspired by the success of the LoRA technique \cite{hu2021lora}. The experimental results in Section~\ref{sec:exp:rank_mapping} illustrate the critical importance of the low-rank mapping function design for our DPG module. Moreover, we apply Dropout \cite{srivastava2014dropout} and LayerNorm \cite{ba2016layer} to this low-rank linear mapping function
\begin{equation}
    Mapping(x) = LayerNorm(LowRankLinear(DropOut(x))),
    \label{eq:mapping_func}
\end{equation}
which further helps to avoid overfitting and stabilize training.

With the help of this specialized mapping function along with a pre-trained neural network, the prompt $P$ is dynamically obtained from the input image $I$ and stage tokens $S_{t, q} := S_{t - (q - 1)} \sim S_{t} $: 
\begin{equation}
    P = Mapping(DPG\mbox{-}Network([S_{t, q}; \hspace{1mm} I])). 
    \label{eq:propmt_generation}
\end{equation}
We reshape $P$ to a size of $N_p \times C \times L$, where $N_p$ is the number of prompts, $C$ is the channel dimension, and $L$ indicates the number of layers where prompts are applied to the backbone model. The generated prompts are inserted across the first $L$ layers in the ViT backbone.
In our DPG approach to prompting, we follow the technical foundations of prompting from the prior state-of-the-art PCL methods \cite{wang2022dualprompt,smith2023coda}. This is because we want to make fair comparison with the previous work, and \textit{our primary focus is on the formation of prompt}. The prompt for the $l$-th layer, denoted as $P_l \in \mathbb{R}^{N \times C}$, is partitioned into $\{P_{l,k}, P_{l,v}\}\in \mathbb{R}^{\frac{N}{2} \times C}$ and is prefixed to the input token embedding for both \textit{key} and \textit{value} in the attention mechanism~\cite{vaswani2017attention}. This can be formulated as: 
\begin{equation}
    h_i = Attention(X_lW_q^i, \hspace{1mm} [P_{k};X]W_k^i, \hspace{1mm} [P_{v};X]W_v^i).
\end{equation}
Here, we omit layer index $l$ for brevity, and $i$ is head index in multi-head self-attention of transformer architecture.

\subsubsection{Joint Dynamic Prompt and Representation Learning.} Differing from prior PCL methods~\cite{wang2022learning, wang2022dualprompt, smith2023coda} that freeze the backbone model (on the right side of the pipeline in Figure~\ref{fig:ours_vs_previous}(a)) during training, we propose our Dynamic Prompt and Representation Learner (DPaRL), to jointly learn the \textit{dynamic prompt generation} with the discriminative representation learning \textit{backbone} (on the right side of Figure~\ref{fig:ours_vs_previous}(b) and Figure~\ref{fig:method}). This approach aims to maximize the capability of entire pipeline to integrate old-stage concepts and new ones to encapsulate diverse semantics that aid generalization to open-world unseen classes and unknown domain shifts. Leveraging parameter-efficient fine-tuning techniques, our DPaRL successfully maximizes accuracy performance and minimizes catastrophic forgetting simultaneously. The overall framework is depicted in Figure \ref{fig:method}. Note that the DPG \textit{encoder} weights (on the left side of the pipeline) for prompt generation are \textit{frozen} in our method, while the \textit{learnable} parameters include stage tokens, weights in the mapping function and representation loss function, as well as the discriminative representation backbone weights.

%% file: sections/4_experiments.tex
\section{Experiments}

In this section, we empirically evaluate the effectiveness of our Dynamic Prompt and Representation Learner (DPaRL) in the established challenging continual open-world visual representation learning.

\subsection{Datasets and Evaluation Metric}

\noindent\textbf{Datasets.} We conducted our training and testing across 4 prominent open-world image retrieval benchmarks: Cars~\cite{krause20133d}, In-Shop~\cite{liu2016deepfashion}, SOP~\cite{oh2016deep} and iNat2018~\cite{van2018inaturalist}. 
We evaluate baselines and our proposed method following our practical continual open-world visual representation learning setting. In this dynamic scenario, classes are introduced incrementally at each learning stage, and the testing phase encompasses classes that have not been encountered during training. Our experiments consist of 10 or 100 stages, with the total training classes approximately distributed uniformly across these stages. 
The detailed data information including number of classes and samples are provided in the \textit{Supplementary Material}.
For inference, we use the entire testing dataset for evaluating all CL stages since there is no overlap between training and testing classes. Thus, our objective is to learn class relations while maintaining generalization across CL stages.

\noindent\textbf{Evaluation Metric.} To evaluate model performance, we compute the Recall@1 metric using the [CLS] token at the last layer of the discriminative representation backbone network. Given our goal to optimize the model's adaptability throughout the CL phases, we report average Recall@1 across all CL stage. Let $R_n$ be the Recall@1 performance at stage $n$,  then we report $R_N = \frac{1}{N_s}\sum_{n=1}^{N_s} R_n$ where $N_s$ is the total number of stages. We also report the forgetting score $F_N$ by following \cite{wang2022dualprompt, smith2023coda} to measure the performance drop on \textit{the same unseen testing dataset} across different training stages, that is, $F_N = \frac{1}{N_s - 1}\sum_{n=2}^{N_s} \max\limits_{m \in \{1,..., n-1\}}(0, R_m - R_n)$. We stress that $R_N$ holds greater importance as it captures both learning capacity and forgetting, while $F_N$ merely offers supplementary context.

\subsection{Baselines and Implementations}

\noindent\textbf{Baselines.} We validate 3 types of representative baselines:  
\begin{itemize}
    \item Rehearsal-based: Experience Replay (ER) \cite{chaudhry2019tiny},
    \item Regularization-based: Learning without Forgetting (LwF) \cite{li2017learning}
    \item Rehearsal-free prompt-based: Learning to Prompt (L2P) \cite{wang2022learning}, DualPrompt \cite{wang2022dualprompt}, and CodaPrompt \cite{smith2023coda}. 
\end{itemize}

\noindent\textbf{Implementation Details.} Consistently across all methods, we adhere to the training setups by following~\cite{smith2023coda}. Specifically, we utilize the Adam optimizer~\cite{kingma2014adam} with a batch size of 128 to train for 20 epochs to ensure comprehensive model convergence. The learning rate is cosine decaying with an initial value of 0.001.

We utilize ArcFace \cite{deng2019arcface} as the training loss function as it is a prevalent choice in open-world representation learning to achieve state-of-the-art performance~\cite{an2023unicom}. For new-stage training, the prototype weights in the loss function of old training stages are appended to the new training stage as initialization. All prototype weights in the loss function, including those corresponding to the classes for old and new stages, are trainable during each training stage.

For established PCL techniques, we maintain the prompt lengths and locations for L2P \cite{wang2022learning}, DualPrompt \cite{wang2022dualprompt}, and CodaPrompt \cite{smith2023coda} as suggested in these works.
For our DPaRL method, in each CL stage, we incorporate a learnable stage token with the first-in-first-out (FIFO) order and a max size of 5 tokens based on insights from Section~\ref{sec:exp:stage_token}. We apply our specialized mapping function in the DPG module to avoid over-fitting and stabilize training. We adopt LoRA~\cite{hu2021lora} as the PEFT method to update backbone weights as guided by Section~\ref{sec:exp:joint_learning}.

\begin{table*}[tb]
\caption{Performance on 4 open-world image retrieval benchmarks with 10 CL stages. We report the averaged \textit{Recall@1} ($R_{N}$) and \textit{Forgetting Score} ($F_{N}$) across all continual learning stages, the results are averaged over 3 runs. We present the \textit{LowerBound} with the zero-shot performance of ImageNet-21k pre-trained ViT-B/16 backbone. \textit{UpperBound} denotes the upper bound performance when trained on the entire dataset at a single stage. All PCL methods, including L2P, DualPrompt, CodaPrompt, and our DPaRL, utilize the same pre-trained encoder, ViT-B/16 on ImageNet-21k, to generate the prompt tokens. 
The numbers in brackets show the \textit{Recall@1 {\color{Green}improvement}} of our DPaRL over the previous best baselines, with the previous best baseline \underline{underlined}. 
}
\label{table:exp:comparison}
   \centering
\small
\resizebox{0.98\textwidth}{!}{%
\begin{tabular}{l|cc|cc|cc|cc|cc}
\hline
Dataset & \multicolumn{2}{c|}{Cars \cite{krause20133d}} & \multicolumn{2}{c|}{In-Shop \cite{liu2016deepfashion}} & \multicolumn{2}{c|}{SOP \cite{oh2016deep}} & \multicolumn{2}{c|}{iNat2018 \cite{van2018inaturalist}}  & \multicolumn{2}{c}{Average}\\
\hline
Metric & $R_{N} (\uparrow)$ & \hspace{-2mm} $F_{N} (\downarrow)$ & $R_{N} (\uparrow)$ & \hspace{-2mm} $F_{N} (\downarrow)$ & $R_{N} (\uparrow)$ & \hspace{-2mm} $F_{N} (\downarrow)$ & $R_{N} (\uparrow)$ & \hspace{-2mm} $F_{N} (\downarrow)$ & $R_{N} (\uparrow)$ & \hspace{-2mm} $F_{N} (\downarrow)$  \\
\hline
\hline
LowerBound & 53.86 & -- & 47.16 & -- 
 & 62.85 & -- &   71.74 & --  & 58.90 & -- \\
UpperBound& 85.96  & --  & 89.28 & -- & 85.21  & --  &   82.68  & --  &  85.78  & -- \\
\hline
ER \cite{chaudhry2019tiny} &\underline{65.38} & 0.49 
 & \underline{82.99}  &  2.48
 & 65.38  & 0.50
 & 36.94  & 18.58
  & 62.67 & 5.51 \\
LwF \cite{li2017learning} &56.53 & 2.91
& 81.52 & 3.39
& 66.18 & 13.20
 & 39.69 & 33.81
  & 60.98 & 13.33 \\
L2P \cite{wang2022learning} & 56.27 & 0.00 
& 67.85 & 0.00
& 76.88 & 0.40
 & 78.95 & 0.00
 & 69.99 & 0.10 \\
DualPrompt \cite{wang2022dualprompt} & 55.54 & 0.00
 & 65.56 & 0.00
& 77.04 & 0.66
 & 78.71 & 0.03
 & 69.21 & 0.17 \\
CodaPrompt \cite{smith2023coda}  & 65.23 & 0.36
&78.61 & 0.00
& \underline{81.62} & 0.01
 & \underline{78.97} & 0.01
 & \underline{76.11} & 0.10 \\
 \hline
\rowcolor{Gray}
\textbf{\ourmethod~(Ours)}  & \hspace{-1mm} 73.22({\color{Green} $\uparrow$7.8})\hspace{-0.5mm} & 0.03 & \hspace{-1mm} 86.28({\color{Green} $\uparrow$3.3})\hspace{-0.75mm} & 0.02 & \hspace{-1mm} 83.69({\color{Green} $\uparrow$2.1})\hspace{-0.75mm} & 0.07 & \hspace{-1mm} 80.02({\color{Green} $\uparrow$1.0})\hspace{-0.75mm} & 0.03 & \hspace{-1mm} 80.80({\color{Green} $\uparrow$4.7})\hspace{-0.75mm} & 0.09 
 \\
\hline
\end{tabular}%
}
\end{table*}

\subsection{Performance Comparison with Prior Arts}
\label{sec:exp:perf_comparison}

Table \ref{table:exp:comparison} shows a comprehensive performance comparison with various CL methods across all four open-world image retrieval tasks. 
We adopt the official codebase of CodaPrompt~\cite{smith2023coda} for the implementation of PCL baselines including L2P \cite{wang2022learning}, DualPrompt \cite{wang2022dualprompt}, and CodaPrompt \cite{smith2023coda}. Additionally, we report the performance of representative rehearsal-based and regularization-based approaches, \ie, ER \cite{chaudhry2019tiny} and LWF \cite{li2017learning}, respectively, for a more comprehensive comparison. 
Note that, for a fair comparison, all methods use the pre-trained ViT-Base on ImageNet-21k as the backbone model, as shown on the right side of Figure~\ref{fig:method}. 

From the results in the Table~\ref{table:exp:comparison}, we observe that the performance trend of PCL-based methods (L2P, DualPrompt, CodaPrompt, our DPaRL) is similar to what was observed in the closed-world image recognition task \cite{smith2023coda}, where PCL methods outperform other continual learning approaches by a significant margin on both Recall@1 metric $R_N$ and Forgetting Score $F_N$\footnote{PCL methods consistently achieve forgetting scores close to zero, indicating that performance continually improves. This trend is also evident in the dynamic change of accuracy curve in the supplementary material, highlighting that PCL methods serve as robust baselines for addressing catastrophic forgetting in open-world tasks.}, aligning with findings from previous research \cite{wang2022learning, wang2022dualprompt, smith2023coda}. This suggests that PCL methods are strong baselines for continual learning on both closed-world image recognition tasks and open-world visual representation learning tasks. Notably, our DPaRL approach consistently outperforms other methods across all datasets, resulting in an average Recall@1 boost of 4.7\% over the leading previous method, CodaPrompt \cite{smith2023coda}.  
This supports our hypothesis about the limitations of the conventional PCL design in open-world scenarios. 

\begin{figure}[t]
\begin{center}
\centering
\def\arraystretch{0.5}
\begin{tabular}{@{\hskip 0.0\linewidth}c@{\hskip 0.0\linewidth}c@{\hskip 0.0\linewidth}c}
\includegraphics[width=0.475\linewidth]{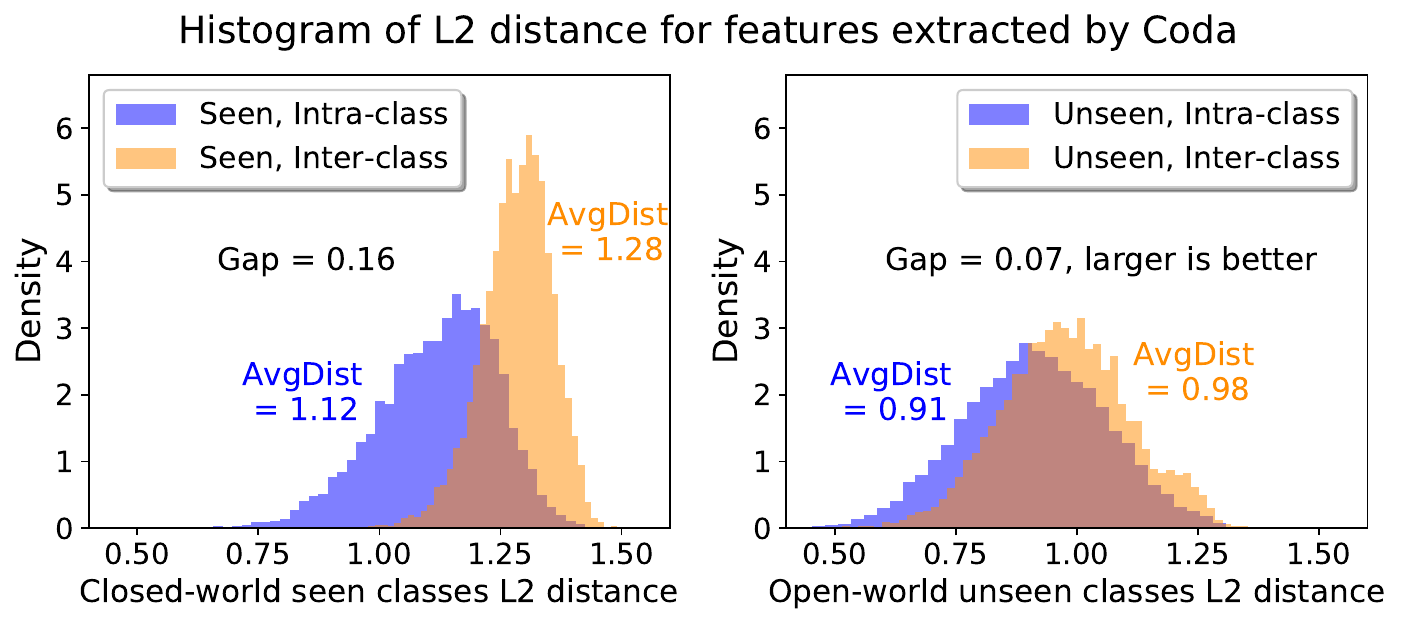} \hspace{1mm} 
\includegraphics[width=0.475\linewidth]{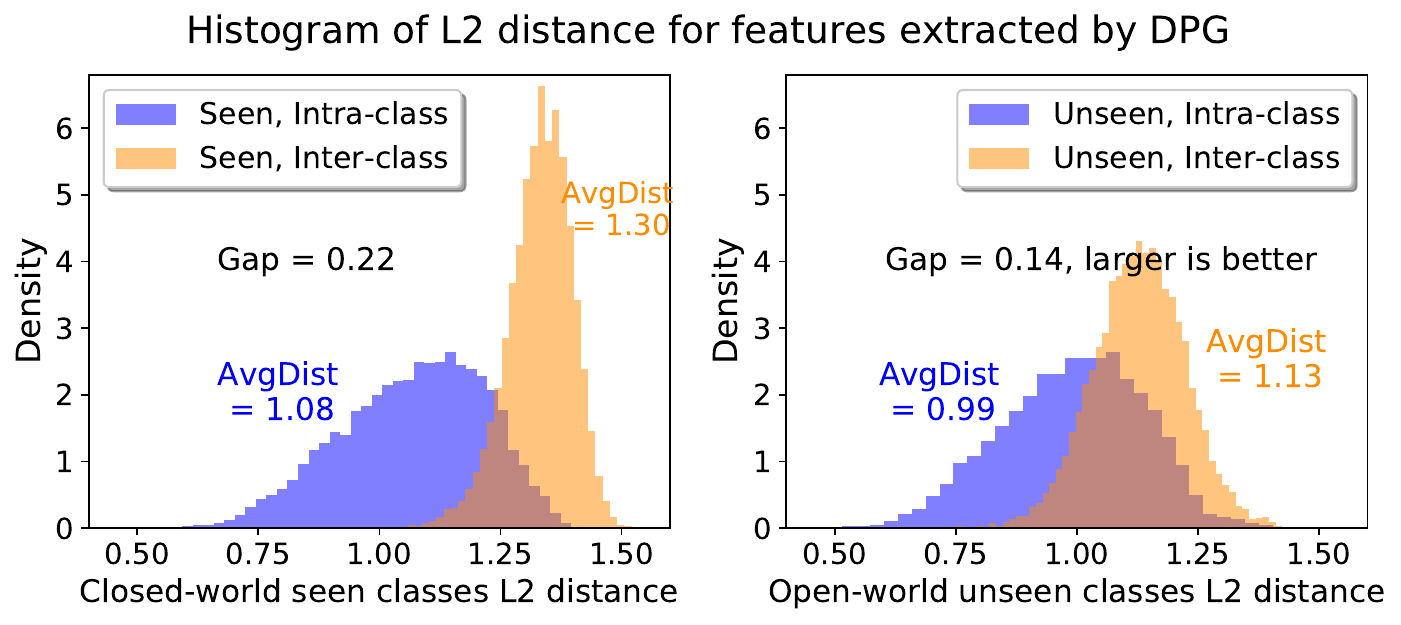}
\end{tabular}
\end{center}
\caption{ 
Histogram of L2 distance between a pair of samples with embedding features extracted by Coda \cite{smith2023coda} (SOTA PCL method) and our DPG (the naive version of DPaRL with freezing backbone weights). Left and right figures are distributions of the seen and unseen classes in training and testing Cars \cite{krause20133d} dataset, respectively. 
Our DPG exhibits enhanced separation between inter- and intra-classes, particularly on open-world unseen test classes.
}
\label{fig:hist_l2_dist}
\end{figure}

In the following, we provide step-by-step guidance on the design choices of our DPaRL. Additional analysis and results are presented in the supplementary material, covering the superior performance of DPaRL in few-shot settings, effectiveness of DPaRL in closed-world benchmarks, dynamic changes in the accuracy curve across CL stages, and robustness of DPaRL against various DPG encoders. Due to space limitations, we refer to the supplementary material for a more comprehensive exploration of these aspects.

\subsection{Effectiveness of Dynamic Prompt Generation} \label{sec:exp:DPG_design}

To solely investigate the effectiveness of our Dynamic Prompt Generation (DPG) design, we freeze the backbone weights similar to prior PCL methods~\cite{wang2022learning, wang2022dualprompt, smith2023coda} in our DPaRL and denoted as the DPG method.

We visualize the distributions of L2 distance between pairwise samples within intra- and inter-classes of CodaPrompt (Coda) \cite{smith2023coda} and our DPG in Figure~\ref{fig:hist_l2_dist} to assess generalizability. A good model should \textit{exhibit distinguishable distributions between intra- and inter-classes for both closed-world seen classes and open-world unseen classes}. Coda, employing a static prompt pool design, achieves a notable separation on closed-world intra- and inter-classes but exhibits a close distribution for open-world intra- and inter-classes, impeding effective generalization to unseen test classes. In contrast, our DPG shows a more distinct separation, with an average gap of 0.14 (twice as large as Coda's gap) between inter- and intra-class distances in testing data. This indicates better generalizability with a dynamic prompt generation mechanism for unseen classes.

Moreover, accuracy results with Recall@1 in Table~\ref{table:exp:stage_token_ablation} and Table~\ref{table:exp:ft_backbone} reveal that our DPG significantly outperforms the best baseline (Coda), with an average of 9.1\% boost in 100 CL stages and 3.2\% boost in 10 CL stages, respectively. This highlights the superiority of Dynamic Prompting Generation design, particularly in practical scenarios involving a large number of stages. 
In the next Section 4.4.1 and Section 4.4.2, we investigate the importance of our design components of stage tokens and mapping function in the DPG network.

\subsubsection{4.4.1 Effectiveness of Stage Tokens \vspace{2mm} \\ } 
\label{sec:exp:stage_token}

\begin{table}[tb]
\caption{
Impact of stage tokens with a large number of CL stages, $N_{stg} = 100$. Recall@1 results on In-Shop, SOP, and iNat2018 datasets are reported. It's worth noting that the Cars dataset is relatively small, preventing its division into 100 stages. DPG is a naive version of our DPaRL method by freezing the backbone weights.
}
\label{table:exp:stage_token_ablation}
   \centering
\small
\scalebox{0.8}{
\begin{tabular}{c|c|c|ccc|c}
\hline
Method &  \# Stage Token & Order  & In-Shop & \hspace{1mm} SOP \hspace{1mm} & \hspace{1mm} iNat \hspace{1mm} &  Average  \\
\hline
L2P~\cite{wang2022learning} & N/A & N/A 
 & 48.39 & 70.39 & 76.82 & 65.20 \\
Dual~\cite{wang2022dualprompt} & N/A & N/A 
 & 52.50 & 73.76 & 76.91 & 67.72\\
\hspace{1mm} Coda~\cite{smith2023coda} \hspace{1mm} & N/A & N/A 
 & \underline{52.96} & \underline{73.92} & \underline{76.98} & \underline{67.95}\\
\hline
DPG & None & N/A & 71.18 & 80.04 & 77.57 &  \hspace{1mm} 76.26 ({\color{Green} $\uparrow$8.31}) \\
\rowcolor{Gray}
DPG & 5 & \textbf{FIFO (default)} & 72.05 & 80.81 & 78.29 &  \hspace{1mm} 77.05 ({\color{Green} $\uparrow$9.10}) \\
DPG & 5 & FILO  & 71.76 & 80.60 & 78.12 & \hspace{1mm} 76.83 ({\color{Green} $\uparrow$8.88}) \\
DPG & 5 & Random  & 72.00 & 80.83 & 78.20 &  \hspace{1mm} 77.01 ({\color{Green} $\uparrow$9.06})  \\
\hline
DPG & 10 & FIFO & 72.15 & 80.81 & 78.28 & \hspace{1mm} 77.08 ({\color{Green} $\uparrow$9.13})\\
DPG &  100 & full-size baseline  & 72.20 & 80.84 & 78.37 & \hspace{1mm} 77.14 ({\color{Green} $\uparrow$9.19}) \\
\hline
\end{tabular}
}
\end{table}

In our DPaRL design with Dynamic Prompt Generation (DPG), we introduce stage tokens to retain information on old training stages, and we impose a max limit on the number of stage tokens ($M_{stg\_tok} = 5$) to control storage and computation costs as the number of stages scales up, e.g., $N_{stg} = 100$. Hence, when the stage number goes beyond the max limit $M_{stg\_tok}$, we replace some of previous stage tokens. \textit{Our hypothesis is that the earlier stage information learned from previous stage tokens can propagate into the later ones.} In this section, we conduct ablations in a practical CL setting with a large number of stages (100 stages) to study the impact of stage tokens. 

\textbf{With and without stage token}. From Table~\ref{table:exp:stage_token_ablation}, DPG with 5 stage tokens achieves a noticeable accuracy improvement over DPG without stage tokens, 77.05 vs. 76.26 on average, indicating the importance of stage token for retaining stage-wise class information in old training stages. 

\textbf{Impact of the order}. We conduct ablations on three different orders: first-in-first-out (\textit{FIFO}), first-in-last-out (\textit{FILO}), and randomly replacing one stage token (\textit{Random}).
Table~\ref{table:exp:stage_token_ablation} reveals that \textit{FIFO} performs the best on average. \textit{FILO} achieves slightly lower accuracy. Nevertheless, the \textit{Random} method achieves performance very close to the best, indicating the robustness of our DPG design to different orders of stage tokens. We hypothesize this due to stage information learned from previous ones effectively propagating to the new stage tokens, regardless of the order or the location where the new stage token is inserted. 

\textbf{Selection of maximum limit $M_{stg\_tok}$}. When the stage number is large ($N_{stg} = 100$), keeping a maximum token number of 5 achieves comparable performances (77.05 vs. 77.14 on average) to that of the full token number baseline with $M_{stg\_tok}=100$. This signifies a highly effective selection with a constant number of stage tokens to achieve high accuracy, as well as good scalability with respect to the stage number. Hence, we use $M_{stg\_tok}=5$ throughout this paper.

\subsubsection{4.4.2 Effectiveness of Mapping Function\vspace{2mm} \\} \label{sec:exp:rank_mapping}

In our DPaRL framework (Figure \ref{fig:method}), as elaborated in Section~\ref{sec:DPG}, we introduce a mapping function to match the size of [CLS] token and prompt tokens. We apply rank-$R$ constraint to avoid overfitting and reduce the memory consumption from the fully-connected layer inside the mapping function. To measure the sensitivity of parameter $R$, we conduct experiments with varying $R$ from 64 to 768. 

The results for both In-Shop and iNat in Table~\ref{table:exp:mapping_func} indicate the existence of an optimal rank with $R_{opt} = 256$ that strikes a balance between accuracy and the number of parameters, saving 2/3 of the total parameters compared to the full rank case. This default rank $R_{opt} = 256$ is subsequently used for all experiments. 

\begin{table}[th]
\caption{
Impact of the rank of weight matrix inside the mapping function on our DPG design. Recall@1 results on In-Shop and iNat2018 with 10 CL stages are reported.
}
\label{table:exp:mapping_func}
   \centering
\small
\resizebox{0.65\textwidth}{!}{%
\begin{tabular}{c|cccccc}
\hline
Rank & 64 & 128 & 256 & 384 & 512 & 768 \\
\hline
\hline
In-Shop & 82.33 & 83.17 & \textbf{84.09} & 83.86 & 82.98 & 71.40 \\
iNat2018 & 79.21 & 79.85 & \textbf{80.02} & 79.74 & 79.63 & 78.98 \\
\hline
\#Parameters & \hspace{1mm}2.0M \hspace{0.5mm} & \hspace{0.5mm} 3.9M \hspace{0.5mm} & \hspace{0.5mm} 7.9M \hspace{0.5mm} & \hspace{0.5mm} 12.0M \hspace{0.5mm} & \hspace{0.5mm} 16.1M \hspace{0.5mm} &  23.5M\\
\hline
\end{tabular}%
}
\end{table}

Note that when the rank reaches 768, representing the full rank of the mapping function essentially serving as a fully-connected layer, a significant performance drop of 12.69\% is observed for the In-Shop dataset in Table~\ref{table:exp:mapping_func}. This suggests that an overparameterized expanding layer results in a severe accuracy degradation for a highly compressed [CLS] token, indicating the importance of our design. Moreover, our design has a robust rank range from 128 to 384, which closely matches the accuracy performance of the peak result.

\subsection{Effectiveness of Dynamic Prompt and Representation Learner}\label{sec:exp:joint_learning}

Previous PCL methods~\cite{wang2022learning, wang2022dualprompt, smith2023coda} freeze backbone weights of the discriminative model to reduce catastrophic forgetting. However, this approach could limit accuracy performance as it doesn't provide the degree of freedom to update the backbone, particularly in open-world setting where data distribution and domain shifts naturally exist. In this section, we explore effective ways to update backbone weights jointly with our dynamic prompt learning.
We explore 3 approaches here: 1) fine-tuning (\textit{FT}) the entire backbone weights, 2) parameter-efficient fine-tuning (PEFT) with AdaptFromer~\cite{chen2022adaptformer} (\textit{Adapter}) or Low-Rank Adaptation~\cite{hu2021lora} weights on the backbone (\textit{LoRA}), and 3) joint prompt generation and representation learning on the backbone weights (\textit{Coda} and \textit{DPaRL, w/ X}).

Table~\ref{table:exp:ft_backbone} shows that vanilla \textit{FT} or fine-tuning with PCL methods not only leads to lower accuracy but also suffers from higher catastrophic forgetting compared to \textit{DPG}. While, \textit{LoRA} gets higher accuracy (+10.4\% on average) and reduces forgetting (-2.3\% on average) compared to vanilla \textit{FT}, as well as outperforms \textit{Adapter}, indicating a more effective way to update the backbone. Still, \textit{LoRA} is less effective than \textit{DPG}, with an average accuracy gap of 1.7\% and a forgetting gap of 0.6\%, indicating the superiority of our DPG design for open-world CL problems. Lastly, our joint Dynamic Prompt and Representation Learner (DPaRL) with LoRA improves average accuracy by 3.2\% and reduces forgetting by 0.5\% compared to LoRA-only baseline. It also boosts accuracy by 1.4\% on average compare to DPG method without updating backbone weights at a negligible cost of 0.09 forgetting raises, indicating the effectiveness of our DPaRL design to maximizing accuracy without causing catastrophic forgetting issues. Therefore, unless stated otherwise, we adopt joint learning with dynamic prompt generation with LoRA as our default DPaRL method in this paper.

\begin{table}[th]
\caption{
Accuracy performance ($R_{N}, \uparrow$ ) and forgetting scores ($F_{N}, \downarrow$) in 10 CL stages with different fine-tuning methods: fine-tuning (FT) the entire backbone weights, PEFT with prompt tuning (Coda and our DPG), Adapter~\cite{chen2022adaptformer}, and LoRA~\cite{hu2021lora}.
}
\label{table:exp:ft_backbone}
   \centering
\small
\resizebox{0.975\textwidth}{!}{%
\begin{tabular}{l|cc|cc|cc|cc|cc}
\hline
Dataset & \multicolumn{2}{c|}{Cars } & \multicolumn{2}{c|}{In-Shop} & \multicolumn{2}{c|}{SOP } & \multicolumn{2}{c|}{iNat2018 }  & \multicolumn{2}{c}{Average} \\
\hline
Metric 
& \hspace{0.5mm} $R_{N} (\uparrow)$ \hspace{0.5mm} & \hspace{0.5mm} $F_{N} (\downarrow)$ \hspace{0.5mm}
& \hspace{0.5mm} $R_{N} (\uparrow)$ \hspace{0.5mm} & \hspace{0.5mm} $F_{N} (\downarrow)$ \hspace{0.5mm}
& \hspace{0.5mm} $R_{N} (\uparrow)$ \hspace{0.5mm} & \hspace{0.5mm} $F_{N} (\downarrow)$ \hspace{0.5mm}
& \hspace{0.5mm} $R_{N} (\uparrow)$ \hspace{0.5mm} & \hspace{0.5mm} $F_{N} (\downarrow)$ \hspace{0.5mm}
& \hspace{0.5mm} $R_{N} (\uparrow)$ \hspace{0.5mm} & \hspace{0.5mm} $F_{N} (\downarrow)$ \hspace{0.5mm}
\\
\hline
Coda~\cite{smith2023coda}, freeze(frz) backbone  & 65.23 & 0.36 & 78.61 & 0.00 & 81.62 & 0.01 & 78.97 & 0.01 & 76.11 & 0.10 \\
\rowcolor{Gray}
\textbf{DPG (Ours), frz backbone}  & 70.62 & \textbf{0.00} 
 & 84.09 & \textbf{0.00}
&  82.69  & \textbf{0.00}
 &  80.02  & \textbf{0.01}
 & 79.36 & \textbf{0.00}
 \\
\hline
FineTune~\cite{smith2023coda} \hspace{-2mm} & 57.57 & 5.04 
 & 82.23  &  0.54
 & 73.45  & 2.23
 & 55.69  & 3.84 
 & 67.24 & 2.91 \\
Adapter~\cite{chen2022adaptformer} & 64.59 & 2.47 & 83.94 & 0.00 & 82.18 & 0.06 & 77.27 & 0.31 & 76.99 & 0.71 \\
LoRA~\cite{hu2021lora} & 67.77 & 0.99
& 83.29 & 0.40
& 81.97 & 0.27
 & 77.47 & 0.79 
 & 77.63 & 0.61 \\
\hline
Coda, w/ FT  & 50.03 & 12.79 & 82.29 & 0.45 & 74.03 & 3.29 & 57.42 & 4.39 & 65.94 & 5.23 \\
Coda, w/ Adapter & 64.39 & 0.86 & 83.56 & 0.45 & 81.98 & 0.05 & 78.56 & 0.06 & 77.12 & 0.36 \\
Coda, w/ LoRA & 69.06 & 1.23 & 82.79 & 0.33 & 82.51 & 0.25 & 79.12 & 0.34 & 78.37 & 0.54 \\
\hline
 \rowcolor{Gray}
 \textbf{DPaRL, w/ FT}  & 65.00 & 4.61 & 82.36 & 0.33 & 74.24 & 2.71 & 56.73 & 4.12 & 69.58 & 2.94 \\
 \rowcolor{Gray}
\textbf{DPaRL, w/ Adapter} & 68.47 & 1.00 & 85.72 & 0.01 & 82.57 & 0.01 & 78.04 & 0.03 & 78.70 & 0.26 \\
 \rowcolor{Gray}
\textbf{DPaRL, w/ LoRA (default)}   & \textbf{73.22} & 0.03 
 & \textbf{86.28} & 0.02
&  \textbf{83.69}  & 0.07
 &  \textbf{80.02}  & 0.22
& \textbf{80.80} & 0.09 \\
\hline
\end{tabular}%
}
\end{table}

\subsection{Performance on Closed-World Setting in Other Domains} \label{sec:closed_world_eval}

In this section, we assess the effectiveness of our DPaRL method in domains other than the established image retrieval benchmarks such as medical image domain. However, there are no available open-world benchmarks so we follow the standard \textit{closed-world} setting in Figure~\ref{fig:problem_setting} (b).
We conduct experiments on MedMNIST~\cite{yang2023medmnist}, which consists of approximately 700K biomedical images. It includes 7 datasets, each with 7$\sim$11 classes. We evenly split the classes into 3 CL stages and report the average top-1 accuracy over each stage.
Results in Table~\ref{table:exp:med_mnist} demonstrate that DPaRL generalizes well to the medical domain, outperforming other methods with an average improvement of 6.50\%. For additional tasks in other domains, please refer to the supplementary material.

\begin{table}[th]
\caption{
Top-1 (\%) on medical data MedMNIST~\cite{yang2023medmnist}. Our DPaRL surpasses other PCL methods by 6.50\% on average. 
}\label{table:exp:med_mnist}
   \centering
\small
\resizebox{0.9\textwidth}{!}{%
\begin{tabular}{c|ccccccc|c}
\hline
 Methods \hspace{0.5mm}  & \hspace{0.5mm} Blood \hspace{0.5mm}	& \hspace{0.5mm} Derma \hspace{0.5mm}	& \hspace{0.5mm} OrganA \hspace{0.5mm} & \hspace{0.5mm} OrganC \hspace{0.5mm} & \hspace{0.5mm} OrganS \hspace{0.5mm} & \hspace{0.5mm} Path \hspace{0.5mm} & \hspace{0.5mm} Tissue \hspace{0.5mm} & \hspace{0.5mm} Average \hspace{0.5mm}  \\
\hline
L2P & 56.03	& 34.00	& 61.12	& 52.87	& 46.25	& 56.39	& 41.29	& 49.71
 \\
Dual & 66.28 & 33.16 & 65.42 & 57.79	& 48.22	& 60.69	& 40.34	& 53.13
 \\
Coda & 57.37 & 33.72 & 65.41 & 65.98	& 46.82	& 79.85	& 32.30	& \underline{54.49}
 \\
\hline
\rowcolor{Gray}
\textbf{DPaRL (Ours)} & \textbf{70.49}	& \textbf{38.71}	& \textbf{70.39}	& \textbf{71.29}	& \textbf{51.23}	& \textbf{81.44}	& \textbf{43.38}	& \textbf{ 60.99({\color{Green} $\uparrow$6.50}) }
\\
\hline
\end{tabular}
}
\end{table}

%% file: sections/5_conclusion.tex
\section{Conclusion}

In this paper, we introduce a practical setting for continual visual representation learning in the open-world. Existing continual learning methods fail to generalize well in this proposed setting. To mitigate this challenge, we propose Dynamic Prompt and Representation Learner (DPaRL), to enhance the adaptability and performance in continual learning for open-world visual recognition. 
Our method consistently demonstrates superior performance across various open-world image retrieval datasets and settings, by dynamically generating prompts from a deep neural network coupled with a dedicated mapping function, and effectively updating the discriminative representation backbone. Notably, it overcomes the intrinsic constraints found in static prompt pool designs, thereby expanding the limits of model generalization in open-world scenarios.


%% file: supp_material.tex
\title{Supplementary Material for \\Open-World Dynamic Prompt and \\Continual Visual Representation Learning}

\titlerunning{DPaRL for Open-World Continual Learning}


\author{Youngeun Kim\inst{1,}\thanks{Equal contribution.}\thanks{Work conducted during an internship at Amazon.}, \hspace{0.1cm} Jun Fang\inst{2,3, *}\thanks{Corresponding author.}, \hspace{0.1cm} Qin Zhang$^2$, \hspace{0.1cm} Zhaowei Cai$^3$, \\ Yantao Shen$^2$, \hspace{0.1cm} 
Rahul Duggal$^2$, \hspace{0.1cm} Dripta S. Raychaudhuri$^2$, \\ Zhuowen Tu$^2$, \hspace{0.1cm} Yifan Xing$^2$, \hspace{0.1cm} Onkar Dabeer$^2$ 
}

\authorrunning{Y.~Kim et al.}

\institute{Yale University
\and
AWS AI Labs \and Amazon AGI
\\ \email{youngeun.kim@yale.edu  }  \email{\{junfa, qzaamz, zhaoweic, yantaos, dugrahul, driptarc, ztu, yifax, onkardab\}@amazon.com}
}

\maketitle

\setcounter{table}{0}
\renewcommand{\thetable}{\Alph{table}}

\setcounter{figure}{0}
\renewcommand{\thefigure}{\Alph{figure}}

\appendix

In this supplementary material, we expand our discussions on Dynamic Prompt and Representation Learner (DPaRL) with additional analysis and experiments. In particular, we discuss the following:
\begin{itemize}
    \item We provide the detailed data information in Section~\ref{supp:sec:data_info} on our established open-world continual representation learning tasks with four benchmarks;
    \item We compare the number of prompt tokens and learnable parameters for Prompt-based Continual Learning (PCL) methods in Section~\ref{supp:sec:num_prompts_paras};
    \item We present the dynamic changes in the accuracy curve across all continual learning stages with various PCL methods in Section~\ref{supp:sec:across_CL_stages};
    \item We study the few-shot performance  in Section~\ref{supp:sec:few_shot} with various PCL methods and demonstrate the superior performance of our DPaRL approach;
    \item We illustrate the effectiveness of our DPaRL in closed-world evaluation benchmarks in Section~\ref{supp:sec:closed_world_eval};
    \item We investigate the influence of various DPG encoders in the PCL methods in Section~\ref{supp:sec:various_DNNs};
    \item We report the empirical training and testing time in Section~\ref{supp:sec:time_compare} for different continual learning methods.
\end{itemize}

\section{Data Information}\label{supp:sec:data_info}

In Table~\ref{table:data_info}, we provide the detailed number of classes and data samples we used for four open-world image retrieval benchmarks: Cars~\cite{krause20133d}, In-Shop~\cite{liu2016deepfashion}, SOP~\cite{oh2016deep} and iNat2018~\cite{van2018inaturalist}. 

The Cars dataset \cite{krause20133d} offers 16,185 images of 196 car classes. Models were trained on the first 98 classes and tested on the subsequent 98. The In-Shop dataset \cite{liu2016deepfashion} provides 72,712 clothing images across 7,986 classes, with the first half utilized for training and the latter half, which is divided into a query and gallery set, for testing. The SOP \cite{oh2016deep} includes 120,053 product images spanning 22,634 classes and 24 superclasses, bifurcated approximately in the middle for training and testing. Lastly, iNat2018 \cite{van2018inaturalist} is a fine-grained image retrieval dataset with 461,939 images featuring a diverse range of animal and plant species, comprising 5,690 training classes and 2,452 testing classes. 

We split the number of classes almost evenly across all 10 or 100 Continual Learning (CL) stages in the training data. Except for the first CL stage, all other stages have an equal number of classes. Detailed number of classes in the first and other stages are listed in the Table~\ref{table:data_info}.  

\begin{table}[h]
\caption{
Data information details for four datasets: Cars, In-Shop, SOP, and iNat2018. We report the total number of images, total number of classes, and the average number of images per class in both training and testing. We split the number of classes almost evenly across all 10 or 100 Continual Learning (CL) stages. Except for the 1st CL stage, all other stages have an equal number of classes.
}
\label{table:data_info}
   \centering
\small
\resizebox{0.85\textwidth}{!}{%
\begin{tabular}{c|c|c|c|c|c}
\hline
\multicolumn{2}{c|}{Dataset Information} & \hspace{1mm} Cars \hspace{1mm}  & \hspace{0.5mm} In-Shop \hspace{0.5mm} & \hspace{1mm} SOP \hspace{1mm}   & \hspace{0.5mm} iNat2018 \\
\hline \hline
\multirow{3}{*}{\begin{tabular}[c]{@{}c@{}} Training\\ Data\end{tabular}} 
& Number of images  & 8,054 & 25,882  & 59,551 & 325,846  \\ 
& Number of classes  & 98    & 3,997   & 11,318 & 5,690    \\
& \# Images per class  & 82.2  & 6.5     & 5.3    & 57.3     \\
\hline
\multirow{2}{*}{\begin{tabular}[c]{@{}c@{}}10 CL Stages\end{tabular}} & \begin{tabular}[|c]{@{}c@{}}\# Cls in 1st stage\end{tabular} & 8     & 397     & 1139   & 569      \\
& \begin{tabular}[|c]{@{}c@{}}\# Cls in other stages\end{tabular} & 10    & 400     & 1131   & 569      \\
\hline 
\multirow{2}{*}{\begin{tabular}[c]{@{}c@{}}100 CL Stages\end{tabular}} & \begin{tabular}[|c]{@{}c@{}}\# Cls in 1st stage\end{tabular} & N/A     & 37     & 131   & 47      \\
& \begin{tabular}[|c]{@{}c@{}} \hspace{0.5mm} \# Cls in other stages \hspace{0.5mm}\end{tabular} & N/A    & 40     & 113   & 57      \\
\hline 
\multirow{3}{*}{\begin{tabular}[c]{@{}c@{}}Testing \\ Data\end{tabular} } & Number of images & 8,131 & 26,830  & 60,502 & 136,093  \\
& Number of classes & 98    & 3,985   & 11,316 & 2,452    \\
& \# Images per class  & 83.0  & 6.7     & 5.3    & 55.5  \\
\hline
\end{tabular}
}
\end{table}

For Cars, In-Shop, and SOP, the class labels in each CL stage are following the order of label indices. In another word, the class labels in the $i$-th CL stage are $\{L_i, L_i + 1, L_i + 2, ..., L_{i+1} - 1\}$, where $L_0 = 0$ and $L_{i + 1} - L_{i}$ is the number of classes in the $i$-th CL stage.  
For the iNat2018 dataset, we split the class labels in each CL stage by randomly shuffling the class indices. This step is important for such large-scale and fine-grained dataset as varying class orders can change the task complexity~\cite{smith2023coda}.

\section{Number of Prompt Tokens and Learnable Parameters}\label{supp:sec:num_prompts_paras}

In this section, we provide a detailed comparison of the number of prompt tokens fed into the discriminative backbone model, as well as the total number of learnable parameters for all PCL methods, inlcuding L2P~\cite{wang2022learning}, DualPrompt~\cite{wang2022dualprompt}, CodaPrompt~\cite{smith2023coda}, and our DPaRL. 

\begin{table}[h]
\caption{Number of prompt tokens and learnable parameters in PCL methods.
}
\label{table:exp:prompt_token_number}
   \centering
\small
\resizebox{0.75\textwidth}{!}{%
\begin{tabular}{c|c|c|c|c}
\hline
\multicolumn{1}{c|}{\multirow{2}{*}{Method}} & \multicolumn{3}{c|}{Prompt Token Information} & \multirow{2}{*}{\begin{tabular}[c|]{@{}c@{}}  Learnable \\ \hspace{0.5mm} Parameters \hspace{0.5mm} \end{tabular}} 
\\ \cline{2-4}
\multicolumn{1}{c|}{} & \multicolumn{1}{c|}{\hspace{0.5mm} Pool Size \hspace{0.5mm}} & \multicolumn{1}{c|}{\hspace{0.5mm} Length \hspace{0.5mm}} & \multicolumn{1}{c|}{\hspace{0.5mm} Depth \hspace{0.5mm}} &     \\ \hline \hline
L2P~\cite{wang2022learning}  & 30  & 20  & 5  & 2.42M \\
Dual~\cite{wang2022dualprompt} & 10  & 26  & 5  & 0.49M \\
Coda~\cite{smith2023coda} & 100 & 8   & 5  & 3.84M\\
\hline
\rowcolor{Gray}
\hspace{0.5mm} \ourmethod~(Ours) \hspace{0.5mm} & \textbf{0}  & 8   & 5  & 8.13M \\ \hline
\end{tabular}
}
\end{table}

Prior PCL methods utilize a static prompt pool to select and combine several prompt tokens (prompt length), which are then fed into the ViT backbone model with multiple layers (prompt depth). Therefore, the learnable parameters come from the construction of the prompt pool and the prompt token formation mechanism. The number of these parameters, listed in Table~\ref{table:exp:prompt_token_number}, ranges from 0.5M to 3.8M, making these methods parameter-efficient.

In contrast, our DPaRL is a dynamic prompt generation method that does not rely on a static prompt pool. To investigate the influence on prompt formation, we follow CodaPrompt~\cite{smith2023coda} by setting the same prompt length ($8$) and prompt depth ($5$). We generate prompt tokens dynamically, resulting in the same prompt token size of $8 \times 768 \times 5$. However, our DPG requires 8.1 million learnable parameters, more than double the size of CodaPrompt. This increase is primarily from the specialized mapping function (7.9M) and the LoRA layer weights (0.07M), indicating a stronger representation power compared to methods relying on static prompt pools.

\section{Performance across Continual Learning Stages} \label{supp:sec:across_CL_stages}

\begin{figure*}[t]
\begin{center}
\centering
\def\arraystretch{0.5}
\begin{tabular}{@{\hskip 0.0\linewidth}c@{\hskip 0.0\linewidth}c@{\hskip 0.0\linewidth}c@{\hskip 0.0\linewidth}c@{\hskip 0.0\linewidth}c}
\includegraphics[width=0.245\linewidth]{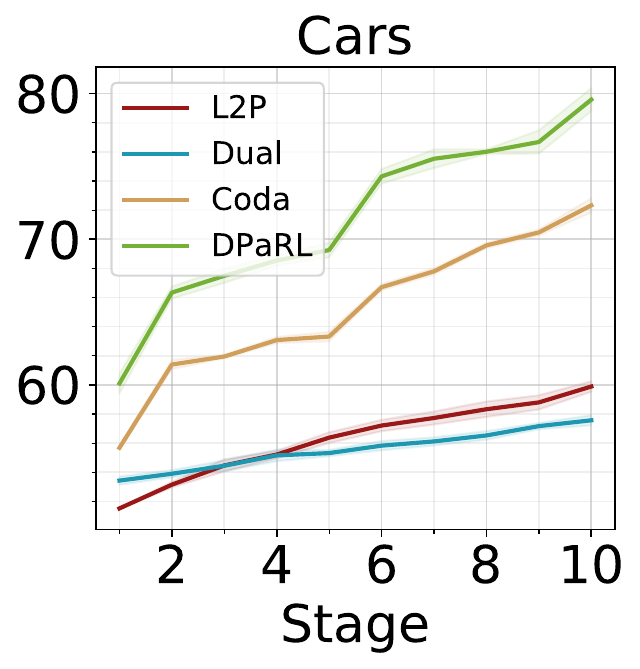}
&
\includegraphics[width=0.245\linewidth]{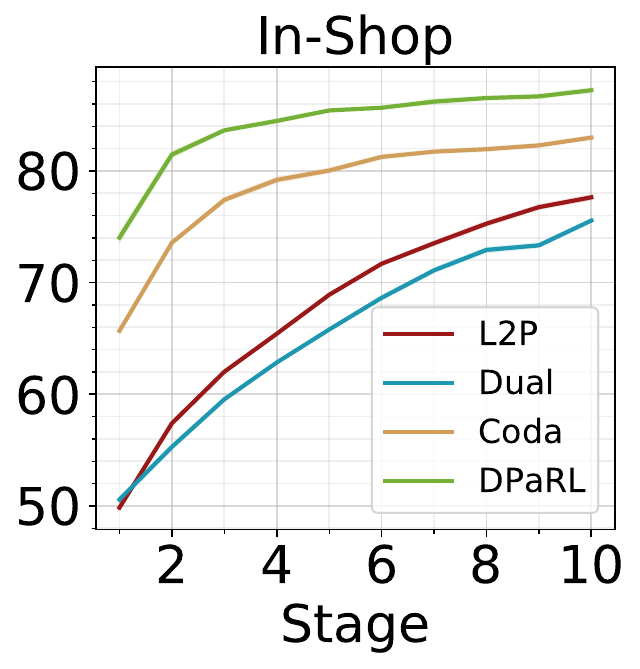}
&
\includegraphics[width=0.245\linewidth]{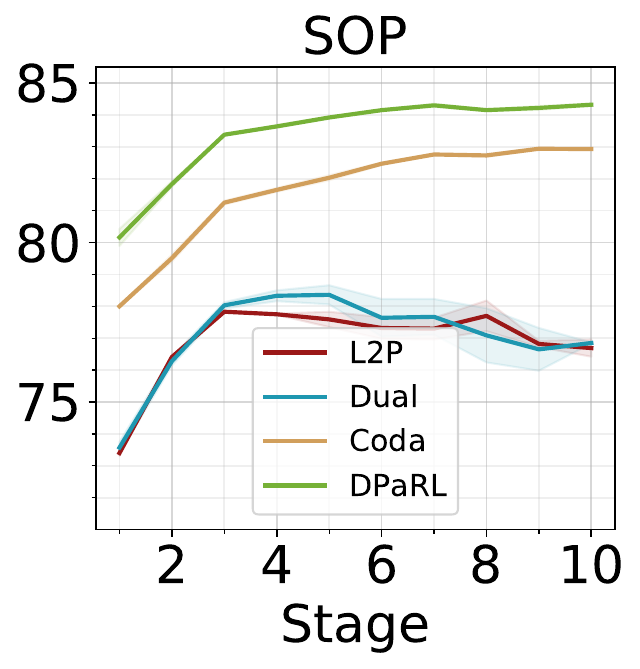}
&
\includegraphics[width=0.245\linewidth]{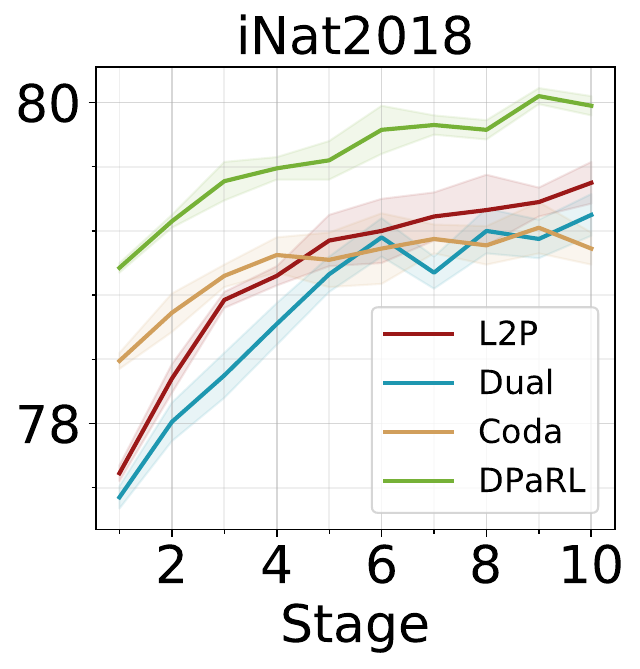}
\end{tabular}
\end{center}
\caption{ 
The change of accuracy in Recall@1 across 10 continual learning stages. Learning to Prompt (L2P)~\cite{wang2022learning}, DualPrompt (Dual)~\cite{wang2022dualprompt}, CodaPrompt (Coda)~\cite{smith2023coda} are static prompt pool-based methods. Our method \ourmethod~is a dynamic prompt generation method. The plots are best viewed in color.
}
\label{fig:exp:performance_trend}
\end{figure*}

In this section, we illustrate the dynamic change of Recall@1 performance throughout the continual learning stages in the Figure \ref{fig:exp:performance_trend} for PCL methods, including Learning to Prompt (L2P)~\cite{wang2022learning}, DualPrompt (Dual)~\cite{wang2022dualprompt}, CodaPrompt (Coda)~\cite{smith2023coda} and our Dynamic Prompt and Representation Learner (\ourmethod). 

Remarkably, from the performance trend, achieving superior performance in initial stages offers a significant advantage, given the unpredictability of when the model might be evaluated in real-world scenarios.
From the figure, we see the common trends: With more continual learning stages and training with more data, Recall@1 performance improves; The methods achieving high Recall@1 in the last stage are likely to show higher Recall@1 in the early stages.

Importantly, our \ourmethod~approach exhibits the capacity to produce generalizable features from the early stages of continual learning, marking a clear advantage over traditional static prompt pool-based techniques.

\begin{figure}[t]
\begin{center}
\centering
\def\arraystretch{0.5}
\begin{tabular}{@{\hskip 0.0\linewidth}c@{\hskip 0.05\linewidth}c@{}c}
\includegraphics[width=0.4\linewidth]{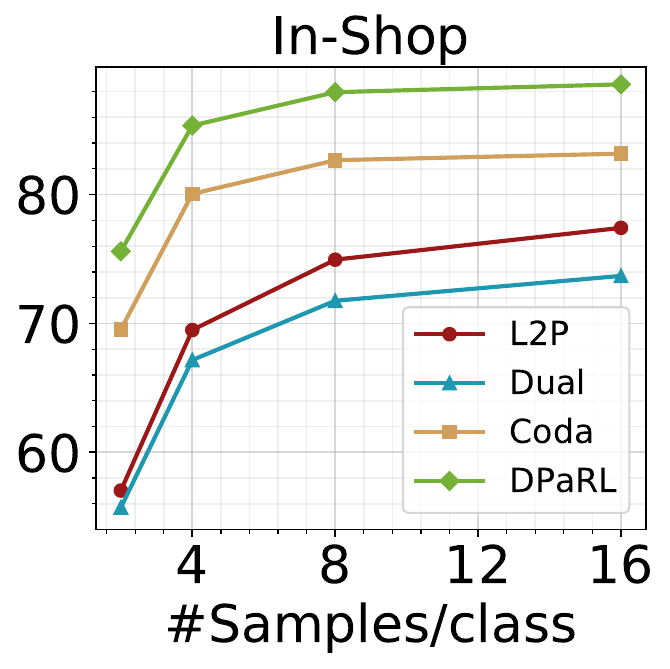} 
&
\includegraphics[width=0.4\linewidth]{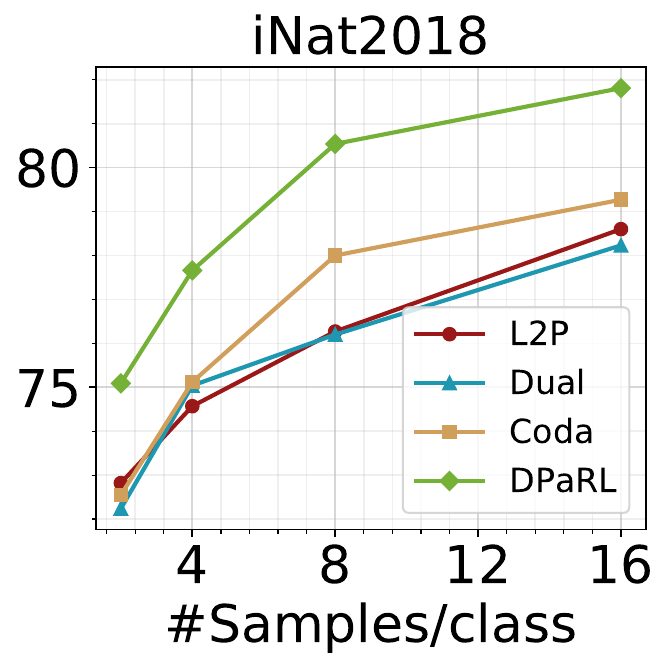}
\end{tabular}
\end{center}
\caption{ 
Accuracy in Recall@1 with respect to the different few-shot continual learning settings with 10 CL stages. The plots are best viewed in color.
}
\label{fig:exp:fewshot}
\end{figure}

\section{Performance in the Few-Shot Setting}\label{supp:sec:few_shot}

In the literature, prompt learning has recently emerged as a prominent technique in the domain of few-shot learning with small number of training samples per class. Consequently, we investigate the efficacy of PCL-based methods within a few-shot continual learning context. 

In Figure \ref{fig:exp:fewshot}, we present the outcomes for the $M$-shot scenarios, where each class is represented by $M$ samples, with $M = 2, 4, 8, 16$, on In-Shop and iNat2018 datasets. It's worth noting that the SOP dataset is inherently designed for few-shot learning, containing approximately 5 samples per class.

The experimental results consistently demonstrate the superior performance of our \ourmethod~methodology across varied few-shot learning scenarios.
Specifically, our method achieves $6.06\%$ and $2.31\%$ higher performance than CodaPrompt on In-Shop and iNat2018 datasets, respectively, with difficult 2-shot learning setting.
This highlights the robust generalization capabilities of our method with dynamic prompts generation and effective representation backbone learning in the few-shot setting.

\section{Performance on Closed-World Benchmarks} \label{supp:sec:closed_world_eval}

In this section, we assess the effectiveness of our method on the closed-world evaluation benchmarks. By following~\cite{wang2022dualprompt, smith2023coda, zhou2023revisiting}, we evaluate our DPaRL on 3 challenging closed-world benchmarks: dual-shift \textit{ImageNet-R} dataset, \textit{DomainNet} and \textit{VTAB} datasets with diverse classes from multiple complex realms.  
Table~\ref{table:exp:comparison_closed_set} presents a comparison to prior methods, with baseline numbers sourced from the original papers. Despite being primarily designed for open-world scenarios, our DPaRL approach demonstrates effectiveness in closed-world benchmarks by outperforming other methods. Particularly, our DPaRL achieves up to 1.47\% of accuracy improvement over the previous best baselines, indicating the superiority of our method.

\begin{table}[h]
\caption{
Accuracy (\%) on closed-world benchmarks. We mark ``--'' if prior work did not report evaluation results on the target dataset.
}
\label{table:exp:comparison_closed_set}
   \centering
\small
\resizebox{0.95\textwidth}{!}{%
\begin{tabular}{c|c|c|c|c}
\hline
Methods & Venue  &  ImageNet-R & DomainNet & VTAB \\
\hline
L2P~\cite{wang2022learning} & CVPR 2022 & 71.66 & 70.54 & 77.11  \\
DualPrompt~\cite{wang2022dualprompt} & ECCV 2022 & 71.32 & 70.73 & 83.36  \\
CodaPrompt~\cite{smith2023coda} & CVPR 2023 & \underline{75.45} & \underline{73.24} & 85.09  \\
ADAM /w Adapter~\cite{zhou2023revisiting} \hspace{0.5mm} & ArXiv 2023 & 72.35 & -- & \underline{85.95}\\
ADP~\cite{tang2023prompt} & ICCV 2023 & 73.27 & -- & -- \\
HiDePrompt~\cite{wang2023hierarchical} & \hspace{0.5mm} NeurIPS 2023 \hspace{0.5mm} & 75.06 & -- & --  \\
\hline
\rowcolor{Gray} 
\textbf{\ourmethod~(Ours)}   &   \textbf{Ours 2024}   &  \hspace{0.5mm} \textbf{76.05} ({\color{Green} $\uparrow$0.60}) \hspace{0.5mm} & \hspace{0.5mm}	\textbf{74.71} ({\color{Green} $\uparrow$1.47}) \hspace{0.5mm} & \hspace{0.5mm} \textbf{86.63} ({\color{Green} $\uparrow$0.68}) \\
\hline
\end{tabular}%
}
\end{table}

\section{Robustness against Various DPG Encoders} \label{supp:sec:various_DNNs}

In the DPG network, we use pre-trained encoders to dynamically generate prompt tokens on the fly. We assess the robustness of our DPG design against various individual encoders, including CLIP~\cite{radford2021learning} and DINO-V2~\cite{oquab2023dinov2}, as well as their combinations. All methods utilize the same discriminative representation backbone model pre-trained on ImageNet-21k, allowing us to solely study the effectiveness of the encoder in generating prompt tokens.

We note that CLIP and DINO-V2 process extensive datasets (>100M) containing rich semantic information, which can provide strong prompt instructors for open-world task. Moreover, none of them rely on labeled data or possess awareness of class information relevant to our open-world tasks. They align with the principles of our open-world setting. 

From the results in Table \ref{table:exp:fm_models}, we observed an average performance enhancement with a single encoder trained on a larger dataset with CLIP or DINO-V2, compared to ImageNet-21k with 14M training samples. Furthermore, no single model attains peak performance across all datasets, indicating the varied suitability of pre-trained encoders for distinct data domains. 

However, we can combine multiple encoders to dynamically generate prompt tokens via concatenation to feed into our DPG network (denoted as DPG++) for harnessing their individual strengths to further enhance the final accuracy.
Here, we limit our design to two encoders, CLIP and DINO-V2, for two reasons: one of these two models can achieve the best performance across these four tasks, and adding more encoders would substantially increase compute and memory requirements. The results of DPG++ in Table \ref{table:exp:fm_models} demonstrate substantial performance improvements of 2.64\% from 79.77\% to 82.41\% on average across all datasets. Moreover, by leveraging DPG++ with our Dynamic Prompt and Representation Learner, denoted as DPaRL++, it achieves another 1.96\% boost to 84.37\%, which is very close to the upper bound of 85.78\%. This emphasizes the robustness of our DPG and DPaRL design in dynamically harnessing diverse pre-trained encoders to enhance open-world visual representation learning capability. It accentuates the practical impact of our method in addressing challenges posed by open-world continual learning problems.

\begin{table}[h]
\caption{Accuracy in Recall@1 with various pre-trained encoders for prompt generation with the same discriminative backbone. DPG++ and DPaRL++ use two encoders to generate and combine dynamic prompts, achieving more advanced performance.
}
\label{table:exp:fm_models}
   \centering
\small
\resizebox{0.9\textwidth}{!}{%
\begin{tabular}{cccccc|c}
\hline
Pre-trained Encoder & Method & Car & In-Shop & SOP & iNat2018 & Average\\
\hline
\hline
ImageNet-21K  \cite{ridnik2021imagenet}  &  \hspace{1mm}   Coda\cite{smith2023coda} \hspace{1mm}    &  65.23 &	78.61	& 81.62	&78.97 & 76.11
\\
\hline
\hspace{-3mm} ImageNet-21K  \cite{ridnik2021imagenet} \hspace{-3mm}   &    DPG    &  70.62 &	84.09	& 82.69	& 80.02  & 79.36 ({\color{Green} $\uparrow$3.25})
     \\
CLIP   \cite{radford2021learning}           &     DPG   &  \underline{71.88} &	82.83	& \underline{82.75} &	80.57  & 79.51 ({\color{Green} $\uparrow$3.40})  \\
DINO-V2   \cite{oquab2023dinov2}           &    DPG    &  70.67	& \underline{84.47} & 	81.77	& \underline{82.17} & 79.77 ({\color{Green} $\uparrow$3.66})
     \\
\hline
\rowcolor{Gray}
 CLIP + DINO-V2             &  \hspace{-3mm} DPG++ \hspace{-3mm}      & 76.63    &  87.10       &  83.46   &  82.45 & 82.41  ({\color{Green} $\uparrow$6.30}) \\
 \hline
\rowcolor{Gray}
 CLIP + DINO-V2             &  \hspace{-2mm} DPaRL++ \hspace{-2mm}      & \textbf{80.25}    &  \textbf{89.23}       &  \textbf{85.44}   &   \textbf{82.59} & \textbf{84.37}  ({\color{Green} $\uparrow$8.26})  \\
\hline
\end{tabular}
}
\end{table}

\section{Training and Testing Time Comparisons}\label{supp:sec:time_compare}

Table~\ref{table:exp:time_compare} outlines the training and testing wall times, measured in minutes (mins), for non-PCL methods, various PCL methods, and our \ourmethod~method. The results highlight that traditional non-PCL methods, such as ER~\cite{chaudhry2019tiny} and LwF~\cite{li2017learning}, exhibit faster training and testing runtimes compared to PCL methods. This discrepancy arises because PCL methods necessitate an additional network for prompt token generation, either from a static prompt pool (L2P, Dual, Coda) or in a dynamic manner (our \ourmethod). 

The primary distinction between our \ourmethod~and other PCL methods lies in the prompt generation mechanism and the parameter-efficient fine-tuning on the backbone model. As discussed in Section~\ref{supp:sec:num_prompts_paras}, our \ourmethod~introduces additional parameters in the dedicated mapping function, stage tokens, and low-rank adaption layers, resulting in more number of learnable parameters compared to other PCL methods. Consequently, our \ourmethod~incurs about 38\% additional training time overhead. However, during inference, our \ourmethod~exhibits comparable runtime to other PCL methods, indicating its deployment advantage with higher accuracy performance. Moreover, we can achieve even higher accuracy by utilizing two pre-trained encoders in our DPG network for dynamic prompt generation, denoted as \ourmethod++, albeit at a higher cost of inference runtime.

\begin{table}[h]
\caption{
Training and testing wall time, in minutes (mins), for different methods on iNat2018 dataset. The training wall time denotes the duration required to complete training across all CL stages. On the other hand, the testing wall time represents the duration for conducting the image retrieval evaluation benchmark on the testing data.
}
\label{table:exp:time_compare}
   \centering
\small
\resizebox{0.8\textwidth}{!}{%
\begin{tabular}{c|c|c|c}
\hline
Method & Method Type & Training  & Testing  \\
\hline
\hline
ER~\cite{chaudhry2019tiny} & \hspace{-2mm} Rehearsal-based  \hspace{-2mm} & 476.6 mins &  7.6 mins \\ 
LwF~\cite{li2017learning} & \hspace{0.5mm} Regularization-based \hspace{0.5mm} & 500.9 mins  & 7.6 mins \\ 
\hline
L2P~\cite{wang2022learning} & Rehearsal-free PCL & 524.8 mins & 8.4 mins \\
Dual~\cite{wang2022dualprompt} & Rehearsal-free PCL & 510.9 mins 
 & 8.4 mins\\
Coda~\cite{smith2023coda}   &  Rehearsal-free PCL & 526.2 mins & 8.5 mins 		
\\ \hline     
\rowcolor{Gray}
\ourmethod~(Ours)   & Rehearsal-free PCL & 727.6 mins    &	8.5 mins 
\\
\rowcolor{Gray}
\hspace{0.5mm} \ourmethod++ (Ours) \hspace{0.5mm}   & Rehearsal-free PCL &  \hspace{0.5mm} 865.5 mins  \hspace{0.5mm}  & \hspace{0.5mm} 10.2 mins 
\\
\hline
\end{tabular}
}
\end{table}


%
%
